\documentclass{article}

\PassOptionsToPackage{numbers}{natbib}

\usepackage[final]{nips}

\usepackage{microtype}
\usepackage{graphicx}
\graphicspath{{./figures/}}
\usepackage{amsfonts}
\usepackage{mathtools}
\usepackage{amsmath}
\usepackage{amsthm}
\usepackage{bbm}
\usepackage{amssymb}
\usepackage{subcaption}
\usepackage{booktabs} 
\usepackage{enumitem}
\usepackage{natbib}
\usepackage[dvipsnames]{xcolor}
\usepackage{color}
\usepackage{wrapfig}
\usepackage{enumitem}
\usepackage{algorithm}
\usepackage{algorithmic}

\newcommand{\ep}{\mathbb{E}}

\newcommand{\fig}[1]{Fig.~\ref{fig:#1}}
\newcommand{\tabl}[1]{Table~\ref{table:#1}}
\newcommand{\eqn}[1]{Eqn.~\eqref{eqn:#1}}
\newcommand{\secref}[1]{Sec.~\ref{sec:#1}}

\title{Multi-objects Generation with Amortized Structural Regularization}

\author{
  Kun Xu, Chongxuan Li, Jun Zhu\footnote{Corresponding author.} , Bo Zhang \\
  Dept. of Comp. Sci. \& Tech., Institute for AI, THBI Lab, BNRist Center, \\
  State Key Lab for Intell. Tech. \& Sys., Tsinghua University, Beijing, China \\
  \{kunxu.thu,chongxuanli1991\}@gmail.com, \{dcszj,dcszb\}@mail.tsinghua.edu.cn
}

\begin{document}

\maketitle

\begin{abstract}
Deep generative models~(DGMs) have shown promise in image generation. However, most of the existing work learn the model by simply optimizing a divergence between the marginal distributions of the model and the data, and often fail to capture the rich structures and relations in multi-object images.
Human knowledge is a critical element to the success of DGMs to infer these structures.
In this paper, we propose the amortized structural regularization~(ASR) framework, which adopts the posterior regularization~(PR) to embed human knowledge into DGMs via a set of structural constraints.
We derive a lower bound of the regularized log-likelihood, which can be jointly optimized with respect to the generative model and recognition model efficiently.
Empirical results show that ASR significantly outperforms the DGM baselines in terms of inference accuracy and sample quality.
\end{abstract}

\vspace{-.2cm}
\section{Introduction}
\vspace{-.2cm}

Deep generative models (DGMs)~\cite{kingma2013auto,oord2016pixel,goodfellow2014generative} have made significant progress in image generation, which largely promotes the downstream applications, especially on unsupervised learning~\cite{chen2016infogan,dilokthanakul2016deep} and semi-supervised learning~\cite{kingma2014semi,chongxuan2017triple}. 
In most of the real world settings, visual data is often presented as a scene with multiple objects with the complicated relationship among them.
However, most of the existing methods~\cite{kingma2013auto,goodfellow2014generative} focus on generating images with a single main object~\cite{karras2017progressive} and lack of a mechanism to capture the underlying structures among objects. It largely impedes DGMs generalizing to complex scene images. How to solve the problem in an unsupervised manner is still largely open.

\begin{wrapfigure}{r}{0.44\columnwidth}
    \centering
    \vspace{-.5cm}
    \begin{subfigure}[t]{0.44\textwidth}
        \includegraphics[width=\columnwidth]{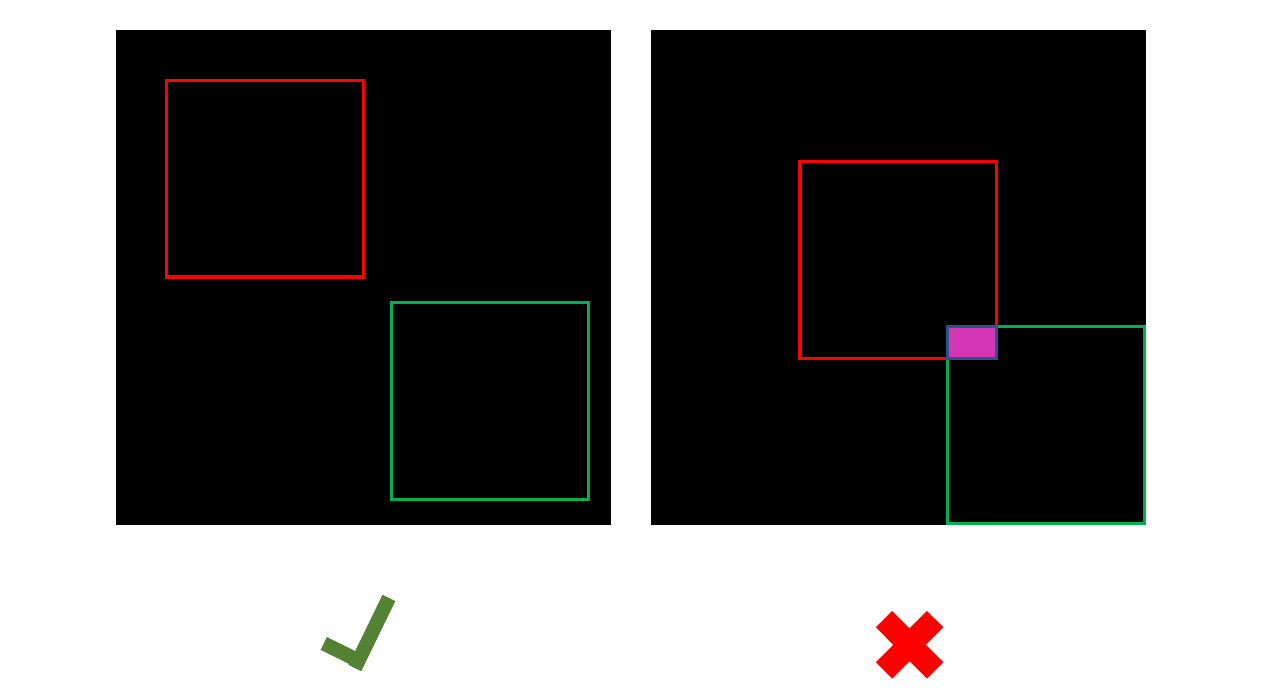}
    \end{subfigure}
    \caption{An illustration of the overlap. The red box denotes the first bounding box, and green denotes the second one. The purple area denotes the overlapping area. The most convenient way to draw non-overlapping bounding boxes is to use rejection sampling. Even for the first bounding box, some locations are not valid. It is difficult to define a proper prior that both tractable and easy to sample.}
    \label{fig:overlap}
    \vspace{-.3cm}
\end{wrapfigure}

Key to the problem is to model the relationships among objects explicitly, which often requires human knowledge to avoid undesirable behavior.
Existing work tends to solve the problem via structured DGMs~\cite{eslami2016attend,li2018graphical}, where a structured prior distribution over latent variables is used to encode the structural information of data and regularize the model behavior.
However, there are two limitations of such methods. 
First, merely maximizing the log-likelihood of such models often fails to capture the structures in the unsupervised manner~\cite{kosiorek2018sequential}. Maximizing the marginal likelihood does not necessarily require the model to capture the reasonable structures as the latent structures are integrated out.
Besides, the optimizing process often stacks in the local optima because of the highly non-linear transformation function defined by neural networks, which also results in undesirable behavior.
Second, it is generally challenging to design a proper prior distribution which is both flexible and computationally tractable.
Consider the case where we want to uniformly sample several $20\times 20$ bounding boxes in a $50\times 50$ image without overlapping. It is difficult to define a tractable prior distribution. An example is shown in Fig.\ref{fig:overlap}. Though it is feasible to set prior distribution to zero through an indicator function when the prior knowledge is violated, it imposes other challenges, including non-convexity and non-differentiability, to the optimization problem. 

In this paper, we propose a flexible amortized structural regularization~(ASR) framework to improve the performance of structured generative models by embedding human knowledge. ASR is based on posterior regularization~(PR) which regularizes structural latent variable models w.r.t. a set of structural constraints. Instead of designing a proper prior, ASR proposes to optimize the log-likelihood of training data along with a regularization term over the posterior distribution. The regularization term can help the model to capture reasonable structures of an image, and to escape the local optima that violate the constraints. 
Specifically, we derive a lower bound of the regularized log-likelihood and introduce a recognition model to approximate the constrained posterior distribution. By slacking the constraints as a penalty term, ASR can be optimized efficiently using gradient-based methods. 

We apply ASR  to the state-of-the-art structured generative models~\cite{eslami2016attend} for the multi-object image generation tasks. Empirical results demonstrate the effectiveness of our proposed method, and both the inference and generative performance are improved under the help of human knowledge.

\vspace{-.2cm}
\section{Preliminary}\label{sec:preliminary}
\vspace{-.2cm}

\subsection{Iterative generative models for multiple objects}
\vspace{-.1cm}

Attend-Infer-Repeat~(AIR)~\cite{eslami2016attend} is a structured latent variable model, which decomposes an image as several objects. The attributes of objects (i.e., appearance, location, and scale) are represented by a set of random variables $z=\{z_{app}, z_{loc}, z_{scale}\}$. 
The generative process starts from sampling the number of objects $n\sim p(n)$, and then $n$ sets of latent variables are sampled independently $z^i\sim p(z)$.
The final image is composed by adding these objects into an empty canvas. Specifically, the joint distribution and its marginal over the observed data can be formulated as follows:
\begin{align}
    p(x, z, n) = p(n)\prod_{i=1:n} p(z^i) p(x|z, n), \:\:\: p(x) = \sum_n \int_z p(x, z, n)  dz. \nonumber
\end{align}
The conditional distribution  $p(x|z, n)$ is usually formulated as a multi-variable Gaussian distribution with mean $\mu=\sum_{i=1:n}f_{dec}(z^i)$, or Bernoulli distribution with $\sum_{i=1:n}f_{dec}(z^i)$ as the probability of 1 for each pixel. $f_{dec}$ is a decoder network which transfers attributes of an object to the image space.

In an unsupervised manner, AIR can infer the number of objects, as well as the latent variables for each object efficiently using amortized variational inference. The latent variables are inferred iteratively and the number of objects $n$ is represented by $z_{pres}$: a $n+1$ binary dimensional vector with $n$ ones followed by a zero. The i-th elements of $z_{pres}$ denotes whether the inference process is terminated or not. Then the inference model can be formulated as follows:
\begin{align}
    q(z, n|x) = q(z_{pres}^{n+1}=0|x, z^{<n})\prod_{i=1:n} q(z^i|x, z^{<i})q(z_{pres}^i=1|x, z^{<i}).
\end{align}
The inference model iteratively infers the latent variables $z^i$ of $i$-th object condition on previous inferred latent variables $z^{< i}$ and the input image $x$ until $z_{pres}^{n+1}=0$.

By explicitly modeling the location and appearance of each object, AIR is capable of modeling an image with structural information, rather than a simple feature vector. It is worth noting that the number of steps $n$, and latent variable $z^i$, are pre-defined and cannot be learned from data.
In the following, we modify the original AIR by introducing a parametric prior to capture the dependency among objects. Details are illustrated in \secref{air-pprior}.

\vspace{-.1cm}
\subsection{Posterior regularization for structured generative model}
\vspace{-.1cm}

Posterior regularization~ (PR)~\cite{ganchev2010posterior,zhu2014bayesian} provides a principled approach to regularize latent variable models with a set of structural constraints.
There are some cases where designing a prior distribution for the prior knowledge is intractable whereas they can be easily presented as a set of constraints~\cite{zhu2014bayesian}.
In these cases, PR is more flexible comparing to designing proper prior distributions.

Specifically, a latent variable model is denoted as $p(X, Z;\theta)=p(Z;\theta)p(X|Z;\theta)$ where $X$ is the training data and $Z$ is the corresponding latent variable. $\theta$ denotes the parameters of $p$, and takes value from $\Theta$, which is generally $\mathbb{R}^{|\Theta|}$ with $|\Theta|$ denotes the dimension of the parameter space.
PR proposes to regularize the posterior distribution to certain constraints under the framework of maximization likelihood estimation (MLE). Generally, the constraints are defined as the expectation of certain statistics $\psi(X,Z)\in \mathbb{R}^d$, and they form a set of valid posterior distribution $Q$ as follows:
\begin{align}
    Q=\{q(Z)| \ep_{q(Z)} \psi(X, Z) \leq \mathbf{0}\}, \nonumber
\end{align}
where $d$ is the number of constraints, and $\mathbf{0}$ is a $d$-dimension zero vector. To regularize the posterior distribution $P(Z|X;\theta)\in Q$, PR propose to add a regularization term $\Omega(p(Z|X; \theta))$ to the MLE objectives. The optimization problem and regularization is given by:
\begin{align}
    \max_{\theta} J(\theta) = \log \int_{Z} p(X, Z; \theta)dZ - \Omega(p(Z|X; \theta)). \label{eqn:pr_obj} \\
    \Omega(p(Z|X;\theta)) = KL(Q||p(Z|X;\theta))=\min_{q\in Q} KL(q(Z)||p(Z|X;\theta)).\label{eqn:pr_Q}
\end{align}
The regularization term is defined as the minimum distance between $Q$ and $p(Z|X;\theta)$ with the distance defined as the KL divergence. When the regularization term is convex, the close-form solution can be found using convex analysis. Therefore, the EM algorithm~\cite{wainwright2008graphical} can be applied to optimizing the regularized likelihood $J(\theta)$~\cite{ganchev2010posterior}. However, EM is largely limited when we extend the PR to DGMs because of the highly non-linearity introduced by neural networks. We therefore propose our method by introducing amortized variational inference to efficiently solve the problem.

\vspace{-.2cm}
\section{Method}\label{sec:method}
\vspace{-.2cm}

\begin{figure}
    \centering
    \vspace{-.1cm}
    \includegraphics[width=\columnwidth]{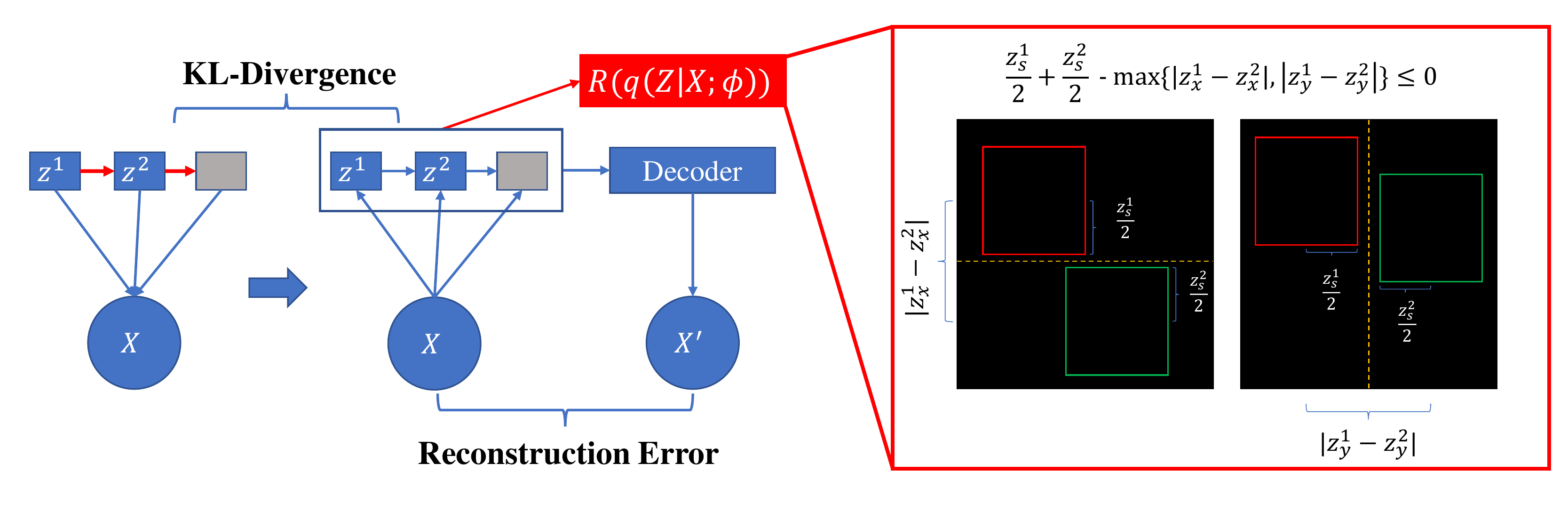}
    \vspace{-.8cm}
    \caption{The proposed framework. The blue arrows denote the generative and inference network in AIR. The red arrows highlight the difference between ASR and AIR. The red arrows in the generative model represent the dependency among the latent variables in the generative model. A regularization term is introduced to regularize the generative model, and we use the overlapping term as an example.}
    \label{fig:framework}
    \vspace{-.5cm}
\end{figure}

In this section, we first define a variant of AIR which uses a parametric prior distribution to capture the dependency of objects. Then we give a formal definition of the amortized structural regularization~(ASR) framework. We mainly follow the notation in \secref{preliminary}, and we abuse the notation when they share the same role in PR and ASR. We illustrate our proposed framework in \fig{framework}.

\vspace{-.1cm}
\subsection{Generative \& inference model}\label{sec:air-pprior}
\vspace{-.1cm}

The prior distribution in vanillan AIR is fixed, and the latent variables of objects are sampled independently. Therefore, the structures, i.e., the attributes and its dependency, cannot be captured by the generative model. 
We propose to modify the generative model by using a learnable prior. Specifically, an auxiliary variable $z_{pres}$ is used to model the number of objects by denoting whether the generation process is terminated at step $t$ (i.e., $z_{pres}^t=0$) or not (i.e., $z_{pres}^t=1$).
Besides, the attributes (i.e., latent variables of each object) are sampled conditioned on previous sampled latent variables. Formally, the joint distribution is defined as follows:
\begin{align}\label{eqn:air-pprior}
& p(x,z,n;\theta) = p(z_{pres}^{n+1} = 0|z^{\leq n};\theta)\left( \prod_{t=1}^{n}p(z^t_{pres}=1|z^{<t};\theta) p(z^{t} | z^{<t};\theta)\right)p(x|z, z_{pres};\theta),  
\end{align}
where the $\theta$ denotes the parameters for both the prior distribution and conditional distribution and we set $z_{pres}^0=1$ and $z^0=\mathbf{0}$. In the following, we omit the $\theta$ for simplicity.
Following AIR, the conditional distribution $p(x|z,z_{pres})$ is defined as $p(x|z, z_{pres})=p(x|\sum_{i=1:n}f_{dec}(z^i))$. We use a recurrent neural network~(RNN)~\cite{graves2013speech} to model the dependency among the latent variables $z$, $z_{pres}$, and use a feed-forward neural network as the decoder to map the latent variables to the image space.

The latent variable $z$ consists of three parts: $z=\{z_{app}, z_{loc}, z_{scale}\}$, which represents the appearance, locations, and scales respectively.
The distribution of $z^{t}$ conditioned on previous $z^{<t}$ is given by:
\begin{align}
    p(z^t|z^{<t};\theta) = p(z_{loc}^t|z^{<t})p(z_{scale}^t|z^{<t},z_{loc}^t)p(z_{app}),\nonumber
\end{align}
where the scale and location is sampled conditionally on previous sampled results, whereas the appearance variables is independently sampled from a simple prior distribution. Since we only consider the spatial relation among them, the dependency among the appearance of objects is ignored.

The inference model is defined mainly following the AIR, which is given by:
\begin{align}\label{eqn:air-pprior-inf}
    q(z, n|x; \phi) = q(z_{pres}^{n+1} = 0|z^{\leq n}, x;\phi)\prod_{t=1}^{n}q(z^i_{pres}=1|z^{<t}, x;\phi) q(z^{t} | z^{<t}, x;\phi),
\end{align}
where the $\phi\in\Phi$ denotes the parameters and $\Phi$ denotes the parameter space of $\phi$. Similar to the generative process, the variational posterior distribution $q(z^{t} | z^{<t}, x;\phi)$ is given by:
\begin{align}
    q(z^{t} | z^{<t}, x) = q(z_{loc}^t|z^{<t})q(z_{scale}^t|z^{<t},z_{loc}^t)q(z_{app}|z_{loc}^t, z_{scale}^t). \nonumber
\end{align}

The generative model defined in \eqn{air-pprior} is powerful enough to capture complex structures. However, directly optimizing the marginal log-likelihood~(or its lower bound) of training data often stacks at the local optima, resulting in the fact that the model fails to capture the structures. Both previous work~\cite{kosiorek2018sequential} and our experimental results present this phenomenon. Details can be found in \secref{exp_num}.    

\vspace{-.1cm}
\subsection{Amortized structural regularization}
\vspace{-.1cm}

In original PR, a set of statistics $\psi$ is used to define the inequality term. In ASR, we generalize the constraints as a functional $F$ that maps a distribution defined over the latent space to $\mathbb{R}^d$, with $d$ denoting the number of constraints.
The resulted valid set $Q$ is given by:
\begin{align}\label{eqn:validset}
Q=\{q(Z)| F(q(Z)) \leq \mathbf{0}\},
\end{align}
where $\mathbf{0}$ is a $d$-dimension zero-vector. 
We only require that the functional $F$ is differentiable w.r.t. $q$ in order to train the DGMs using gradient-based methods efficiently.

Motivated by PR, ASR regularizes the posterior distribution $P(Z|X;\theta)$ within the valid set $Q$, by minimizing a regularization term $\Omega(p(Z|X; \theta))$ along with maximizing the likelihood of training data. The objective function is given by:
\begin{align}\label{eqn:asr_obj_original}
\max_{\theta} J(\theta) = \log \int_{Z} p(X, Z; \theta) dZ - \Omega(p(Z|X; \theta)).
\end{align}

The definition of the regularization term $\Omega$ follows PR as in \eqn{pr_Q}. Note that $KL(q||p(Z|X;\theta)) \geq \Omega(p(Z|X;\theta))$ for all $q(Z)\in Q$. It enables us to obtain a lower bound of $J(\theta)$ by substituting  $\Omega(p(Z|X;\theta))$ to $KL(q(Z)||p(Z|X;\theta))$, which is given by:
\begin{align}
J(\theta) \geq \log \int_{Z} p(X, Z; \theta) dZ - KL(q(Z)||p(Z|X;\theta)) = J^\prime(\theta, q).
\end{align}
Follow the variational inference, the lower bound $J^\prime$ can be formulated as the evidence lower bound~(ELBO), and Problem~\eqref{eqn:asr_obj_original} is converted as a constrained optimization problem as follows:
\begin{align}
    \max_{\theta, q\in Q} &J^\prime(\theta, q) = \ep_{q}\log \frac{p(X,Z;\theta)}{q(Z)}. \nonumber
\end{align}

Motivated by amortized variational inference~\cite{kingma2013auto}, we introduce a recognition model $q(Z|X;\phi)$ to approximate the variational distribution $q$ where $\phi$ denotes the parameters of the recognition model.
Therefore, the lower bound can be optimized w.r.t. $\theta$ and $\phi$ jointly, which is given by:
\begin{align}\label{eqn:ASR_obj}
\max_{\theta\in\Theta, \phi\in\Phi, q(Z|X;\phi)\in Q} \ep_{q(Z|X;\phi)} \log\frac{p(X, Z;\theta)}{q(Z|X;\phi)}.
\end{align}
We abuse the notation $J^\prime(\theta, \phi)$ to denote the amortized version of the lower bound.

Problem~\eqref{eqn:ASR_obj} is a constrained optimization problem.
In order to efficiently solve Problem~\eqref{eqn:ASR_obj}, we propose to slack the constraints as a penalty, and add it to the objective function $J^\prime(\theta, \phi)$ as:
\begin{align}
\max_{\theta\in\Theta, \phi\in\Phi} J^\prime(\theta, \phi) - R(q(Z|X;\phi)),
\end{align} 
where $R(q) = \sum_{i=1:d}\lambda_i\max\{F_i(q), 0\}$, and $\lambda_i$ is the coefficient for the $i$-th constraint of $F(q)$ and acts as hyper-parameters. The training procedure is described in Appendix.

It is worth noting that we implicitly add another regularization to the generative model when defining $q$ using a parametric model: the posterior distribution $p(Z|X;\theta)$ can be represented by $q(Z|X;\phi)$. This regularization term has the same effect as in VAE~\cite{kingma2013auto,shu2018amortized}, which is introduced to make the optimization process more efficient. In contrast, it is the penalty term $R(q(Z|X;\phi))$ that embeds human knowledge into DGMs and regularizes DGMs for desirable behavior. 

\vspace{-.2cm}
\section{Application on multi-object generation}
\vspace{-.2cm}

In the following, we give two examples of applying ASR to image generation with multiple objects. In this section, we mainly focus on regularizing on the number of objects, and the spatial relationships among them. Therefore, the regularization constraints $F$ are defined over $z_{pres}$, $z_{loc}$, and $z_{scale}$. 

\vspace{-.1cm}
\subsection{ASR regularization on the number of objects}
\vspace{-.1cm}

In this setting, we consider the case where each image contains a certain number of objects. For example, each image has either 2 or 4 objects, and images of each number of objects appear of the same frequency. We define the possible numbers of objects as $L\subsetneq [K]$, where $[K]=\{0,1,\cdots,K-1\}$ is the set of all non-negative integer less than K, and $K$ is the largest number of objects we consider. 
Since we use $z_{pres}$ to denote the number of objects, an image $x$ with $n$ objects is equivalent to the corresponding latent variable $z_{pres}|x=u_n$ with probability one, where $u_n$ is a $n+1$ dimension binary vector with $n$ ones followed by a zero. We further denote $q_i$ as $q_i(z_{pres}=u_j)=\mathbbm{1}(i==j)$, where $\mathbbm{1}$ is the indicator function. The valid posterior is given by $V_{z_{pres}}=\{q_i\}_{i\in L}$.
According to ASR, we regularize our variational posterior $q(Z|X;\phi)$ in the valid posterior set $V_{z_{pres}}$. Besides, we also regularize the marginal distribution to $q_{uni}(z)=\frac{1}{|L|}\sum_{i\in L} q_i$, which is a uniform distribution over $V_{z_{pres}}$.
The valid posterior set is given by:
\begin{align}
    Q^{num} = \{q(Z|X)| q(Z|X) \in V_{z_{pres}}, \ep_{p(X)}q(Z|X) = q_{uni}(Z)\}. \nonumber
\end{align}

As the constraints are defined in the equality form, and we reformulate it in the inequality form, and the regularization term $R_{num}$ are given by:
\begin{align}
    Q^{num} = \{q(Z|X)| \min_{q_i \in V_{z_{pres}}} KL(q_i||q(Z|X)) \leq 0, KL(q_{uni}(Z)||\ep_{p(X)} q(Z|X)) \leq 0\}, \nonumber \\
    R^{num}(q(Z|X)) = \lambda_1^{num} \min_{q_i\in Q_{num}} KL(q_i||q(Z|X)) + \lambda_2^{num} KL(q_{u}(Z)||\ep_{p(X)} q(Z|X)). \nonumber 
\end{align}
The $\lambda_1^{num}$ and $\lambda_2^{num}$ are the hyper-parameters to balance the penalty term and the log-likelihood.

\vspace{-.1cm}
\subsection{ASR regularization on overlap}
\vspace{-.1cm}

In this setting, we focus on the overlap problem, and we introduce several regularization terms to reduce the overlap among objects, which is defined over the location of bounding boxes.
The location of a bounding box is determined by its center $z_{loc}=(z_{x}, z_{y})$, and scale $z_{scale}$, and the functional $F^{o}$ is defined over these latent variables.

The first set of regularization terms directly penalize the overlap. Given the centers and scales of the i-th and j-th bounding box, they are not overlapped if and only if both of the following constraints are satisfied:
$\frac{z_{scale}^i+z_{scale}^j}{2} - |z_{x}^i-z_{x}^j| \leq 0, \frac{z_{scale}^i+z_{scale}^j}{2} - |z_{y}^i-z_{y}^j| \leq 0.$ These constraints have a straightforward explanation and is illustrated in \fig{framework}.

In the following, we denote $\ell(x) = \max\{x, 0\}$ for simplicity, and we define the functional $F^{o}$ as:
\begin{align}
    F_1^{o}(q)=\ep_{q(z)}\sum_{i,j<n,i\neq j}\ell(\frac{z_{scale}^i+z_{scale}^j}{2} - \max\{|z_{x}^i-z_{x}^j|, |z_{y}^i - z_{y}^j|\}) \leq 0,  \nonumber 
\end{align}
which regularizes each pair of the bounding boxes to reduce overlapping.

Simply regularizing the overlap by minimizing $F_1$ usually results in the fact that the inferred bounding boxes are of different size: a big bounding box that covers the whole image, and several bounding boxes of extremely small size that lie beside the boundary of the image, or out of the image. To overcome this issue, we add another two regularization terms, where the first one regularize the bounding boxes stay within the image, and the second regularize the bounding boxes are of same size. The first set of regularization term are formulated as the following four constraints:
\begin{align}
    F_2^{o}(q)=\ep_{q(z)} \sum_{i=1:n}\ell(\frac{z_{scale}^i}{2}-z_{x}^i) \leq 0, \:\: F_3^{o}(q)=\ep_{q(z)}\sum_{i=1:n}\ell(z_{x}^i + \frac{z_{scale}^i}{2} -S) \leq 0, \nonumber \\
    F_4^{o}(q)=\ep_{q(z)}\sum_{i=1:n}\ell(\frac{z_{scale}^i}{2} - z_{y}^i) \leq 0, \:\:  F_5^{o}(q)=\ep_{q(z)}\sum_{i=1:n}\ell(z_{y}^i + \frac{z_{scale}^i}{2} -S) \leq 0, \nonumber
\end{align}
and the second set of regularization terms are given by:
\begin{align}
    F_6^{o}(q)=\ep_{q(z)} \sum_{i=1:n}\ell(c_{min}-z_{scale}^i) + \ell( z_{scale}^i-c_{max})\leq 0, \nonumber \\
    F_7^{o}(q)=\ep_{q(z)}\sum_{i,j<n} \ell(|z_{scale}^i-z_{scale}^j|-\epsilon) \leq 0 , \nonumber
\end{align}
where $S$ denotes the size of the final image, $c_{min}$, $c_{max}$ denotes the possible minimum/maximum size of an object, $\epsilon$ denotes the perturbation of the size for objects. Therefore, the regularization for reducing overlapping is given by:
\begin{align}
    R^{o}(q) = \sum_{i=1:7} \lambda_i^{o}F_i^{o}(q).
\end{align}

\vspace{-.8cm}
\section{Related work}
\vspace{-.2cm}

Recently, several work~\cite{eslami2016attend,greff2019multi,johnson2018image,xu2018deep,li2018graphical} introduces structural information to deep generative models. \citet{eslami2016attend} propose the Attend-Infer-Repeat~(AIR), which defines an iterative generative process to compose an image with multiple objects. \citet{greff2019multi} further generalize this method to more complicated images, by jointly modeling the background and objects using masks. \citet{li2018graphical} use graphical networks to model the latent structures of an image, and generalize probabilistic graphical models to the context of implicit generative models.
\citet{johnson2018image} introduce the scene graph as conditional information to generate scene images. \citet{xu2018deep} use the and-or graph to capture the latent structures and use a refinement network to map the latent structures to the image space.

To embed prior knowledge into structured generative models, posterior regularization~(PR)~\cite{ganchev2010posterior} provides a flexible framework to regularize model w.r.t. a set of structural constraints. \citet{zhu2014bayesian} generalize this framework to the Bayesian inference and apply it in the non-parametric setting. \citet{shu2018amortized} introduce to regularize the smoothness of the inference model to improve the generalization on both inference and generation and refer it as amortized inference regularization.
\citet{li2015max} propose to regularize the latent space of a latent variable model with large-margin in the context of amortized variational inference, which can also be considered as a special case of PR. \citet{bilen2014weakly} apply PR to the object detection in a discriminative manner and improve the detection accuracy.

\vspace{-.2cm}
\section{Experiments}
\vspace{-.2cm}

In this section, we present the empirical results of ASR on two dataset: Multi-MNIST~\cite{eslami2016attend} and Multi-Sprites~\cite{greff2019multi}, which are the multi-object version of MNIST~\cite{lecun1998gradient} and dSprites~\cite{higgins2017beta}. We use AIR-pPrior to denote the variants of AIR proposed in this paper, and AIR-ASR to denote the regularized AIR-pPrior using ASR. 

We implement our model using TenworFlow~\cite{abadi2016tensorflow} library. In our experiments, the RNNs in both the generative model and recognition model are LSTM~\cite{hochreiter1997long} with 256 hidden units. A variational auto-encoder~\cite{kingma2013auto} is used to encode and decode the appearance latent variables, and both the encoder and decoder are implemented as a two-layer MLP with 512 and 256 units. We use the Adam optimizer~\cite{kingma2014adam} with learning rate as $0.001$, $\beta_1=0.9$, and $\beta_2=0.999$. We train models with 300 epochs with batch size as 64. Our code is attached in the supplementary materials for reproducing.

In this paper, we use four metrics for quantitative evaluation: negative ELBO~(nELBO), squared error~(SE), inference accuracy~(ACC) and mean intersection over union~(mIoU). The nELBO is an upper bound of negative log-likelihood, where a lower value indicates a better approximation of data distribution.
The SE is the squared error between the original image and its reconstruction, and it is summed over pixels.
The ACC is defined as $\mathbbm{1}(num_{inf}==num_{gt})$, $num_{inf}$ and $num_{gt}$ are the number of objects inferred by the recognition model and ground truth respectively. This evaluation metric demonstrates whether the inference model can correctly infer the exact number of objects in an image. Besides, we also use another evaluation metric mIoU to evaluate the accuracy of inferred location for each objects. The mIoU of a single image is defined as $\max_{\pi}\sum_{i=1:min\{num_{inf}, num_{gt}\}} IoU(z^{\pi_i}, gt^i)/\max\{num_{inf}, num_{gt}\}$, where $\pi$ is a permutation of $\{1,2,\cdots, num_{inf}\}$ and $gt^i$ is the ground truth location for the i-th object.

\begin{figure*}
    \centering
    \begin{subfigure}[t]{0.3\textwidth}
        \includegraphics[width=\textwidth]{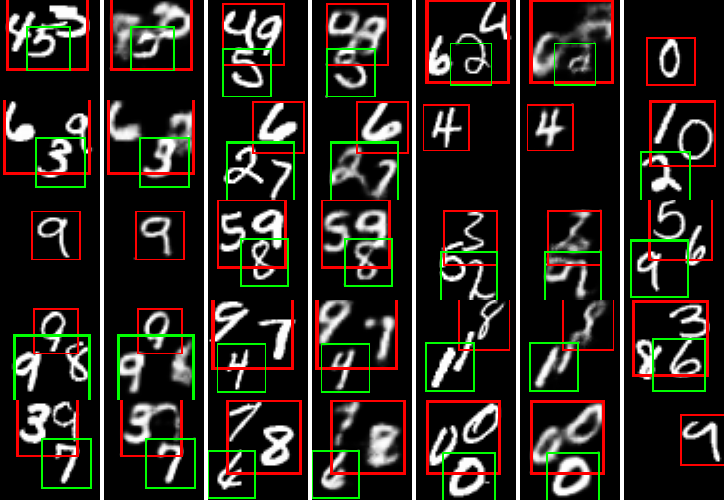}
        \caption{The reconstruction of AIR-13.}
    \end{subfigure}
    \hfill
    \begin{subfigure}[t]{0.3\textwidth}
        \includegraphics[width=\textwidth]{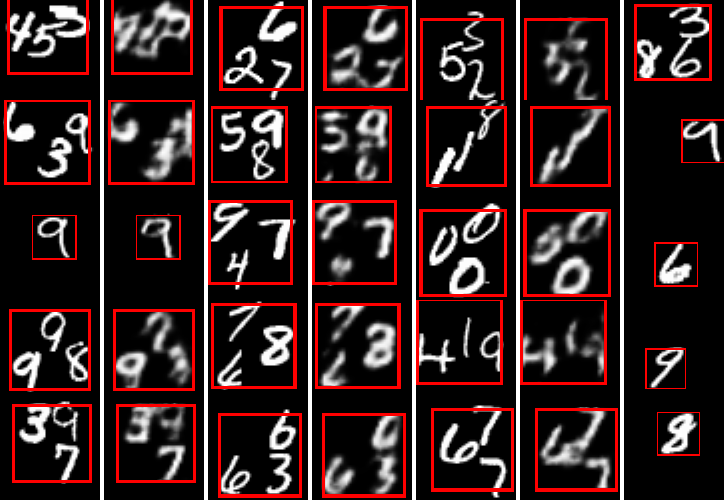}
        \caption{The reconstruction of AIR-pPrior-13.}
    \end{subfigure}
    \hfill
    \begin{subfigure}[t]{0.3\textwidth}
        \includegraphics[width=\textwidth]{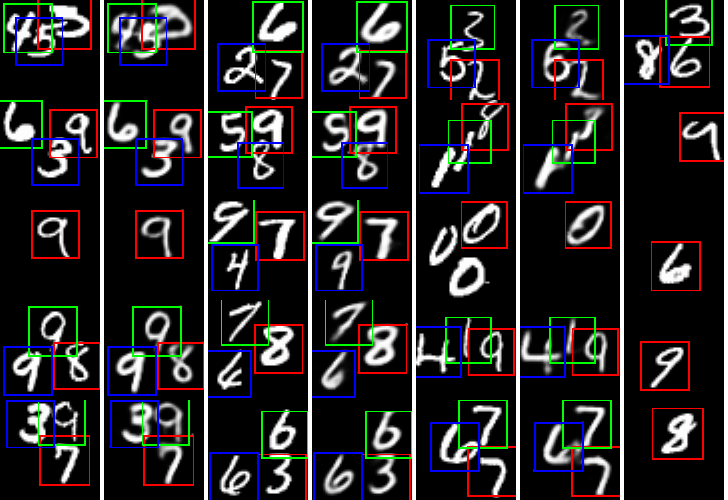}
        \caption{The reconstruction of AIR-ASR-13.}
    \end{subfigure}
    \vspace{-.2cm}
    \caption{The reconstruction results of Multi-MNIST on 1 or 3 objects.}
    \label{fig:num-13}
    \vspace{-.4cm}
\end{figure*}

\begin{table}[htbp]
    \vspace{-.4cm}
    \centering
    \small{
    \caption{Results on regularization on the number of objects. The numbers followed the model name denotes the possible number of objects for a certain image. Results are averaged over 3 runs.}
    \begin{tabular}{|l|cccc|}
        \hline
        Methods      & \multicolumn{1}{c}{nELBO} & \multicolumn{1}{c}{ACC} & \multicolumn{1}{c}{SE} & \multicolumn{1}{c|}{mIoU} \\
        \hline
        AIR-13   &    $404.41\pm 4.58$   &   $0.81\pm0.23$    &   $31.94\pm4.68$   &    $\mathbf{0.61}\pm0.13$     \\
        AIR-pPrior-13 &   $405.21\pm 1.17$    &   $0.48\pm 0.00$    &    $49.42\pm 0.24$   &   $0.43\pm 0.01$      \\
        AIR-ASR-13 &   $\mathbf{360.20}\pm 19.67$    &    $\mathbf{0.96}\pm 0.00$   &  $\mathbf{28.84}\pm 1.11$     &   $\mathbf{0.61}\pm 0.00$    \\
        \hline
        AIR-14   &   $543.44\pm54.71$   &     $0.48\pm0.03$  &  $52.77\pm4.92$     &  $0.43\pm0.07$     \\
        AIR-pPrior-14 &   $519.06\pm 5.47$    &   $0.50\pm 0.00$    &       $68.72\pm 0.55$ &   $0.43\pm 0.00$  \\
        AIR-ASR-14 &   $\mathbf{441.54}\pm 30.97$    &    $\mathbf{0.96}\pm 0.01$   &    $\mathbf{41.05}\pm 7.11$   &  $\mathbf{0.55}\pm 0.08$  \\
        \hline
        AIR-24   &   $639.49\pm23.13$    &   $0.55\pm0.09$    &   $57.69\pm4.88$    &    $0.46\pm0.06$ \\
        AIR-pPrior-24 &   $643.28\pm 8.67$    &   $0.00 \pm 0.00$    &   $83.35\pm 0.44$    &   $0.10\pm 0.00$ \\
        AIR-ASR-24 &   $\mathbf{495.73}\pm 35.80$    &  $\mathbf{0.98}\pm 0.01$     &   $\mathbf{48.54}\pm 5.60$    &  $\mathbf{0.54}\pm 0.08$  \\
        \hline
    \end{tabular}%
    \label{table:num_obj}}
    \vspace{-.3cm}
\end{table}%

\vspace{-.2cm}
\subsection{ASR regularization on the number of objects}\label{sec:exp_num}
\vspace{-.1cm}

\begin{figure}
    \centering
    \begin{subfigure}[t]{0.3\textwidth}
       \includegraphics[width=\textwidth]{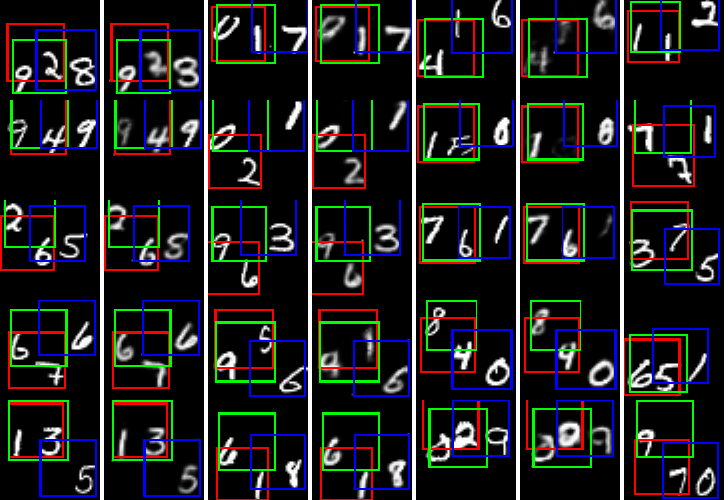}
        \caption{The reconstruction of AIR-3.}
    \end{subfigure}
    \hfill
    \begin{subfigure}[t]{0.3\textwidth}
        \includegraphics[width=\textwidth]{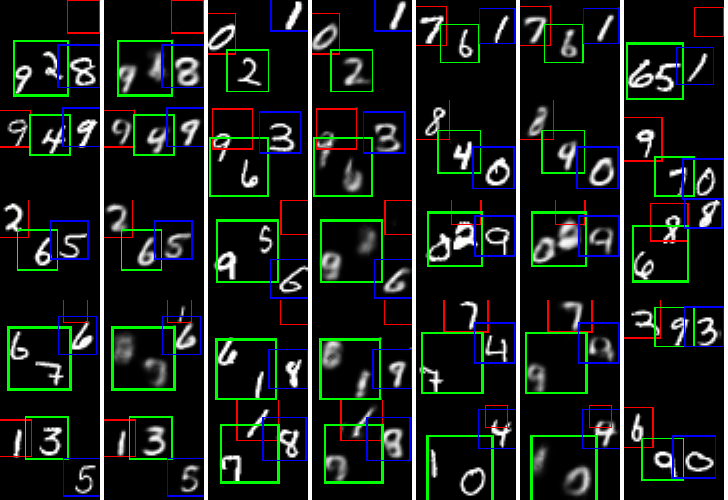}
        \caption{The reconstruction of AIR-pPrior-3.}
    \end{subfigure}
    \hfill
    \begin{subfigure}[t]{0.3\textwidth}
        \includegraphics[width=\textwidth]{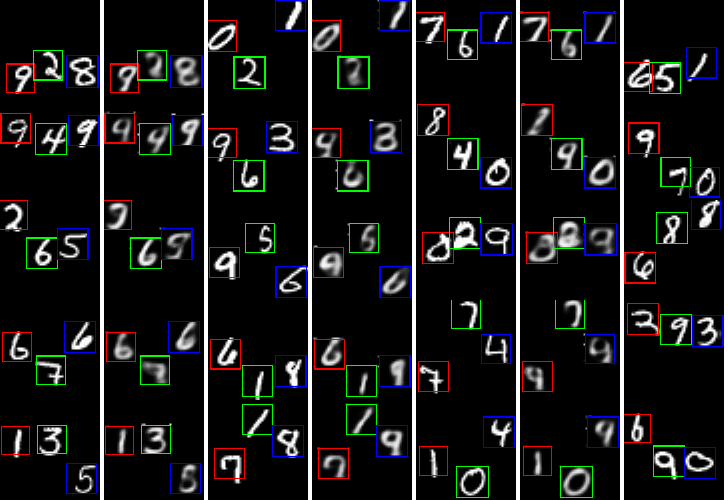}
        \caption{The reconstruction of AIR-ASR-3.}
    \end{subfigure}
    \vspace{-.2cm}
    \caption{The reconstruction results of Multi-MNIST on 3 objects. There is no overlapping among objects in the training data. ASR can successfully infer the underlying structures, and improve the reconstruction results.}
    \label{fig:overlap_mnist_inf}
    \vspace{-.3cm}
\end{figure}

When regularizing on the number of objects, we consider three settings on Multi-MNIST: 1  or 3 objects, 1 or 4 objects, and 2 or 4 objects. 40000 training samples are synthesized where 20000 images for each number of objects. 2000 images are used as the test data to evaluate the performance for inference. In this setting, we evaluate our methods with $\lambda_1^{num}, \lambda_2^{num}\in \{1, 10, 100\}$, and we finally set $\lambda^{num}_1=10$ and $\lambda_2^{num}=100$.

As illustrated in \fig{num-13}, AIR-pPrior simply treats the whole image as a single object, and fails to identify the objects in an image. With a powerful decoder network, the generative model tends to ignore the latent structures and stacks in the local optima. The ASR can successfully regularize the model towards proper behavior and help it escape the local optima. In AIR-ASR, the inference model can successfully identify each object, and the generative model learns the underlying structures.
The original AIR has a better performance compared to AIR-pPrior, as the prior distribution can partly regularize the generative model. However, the original AIR still treats two objects close to each other as one object. The performance of these three models on the other two settings shares the same property, i.e., original AIR tends to merge objects and AIR-pPrior stacks at a local optimum. The other reconstruct results are illustrated in the Appendix.

\tabl{num_obj} presents the quantitative results. AIR-ASR outperforms its baseline and the original AIR on all the evaluation metrics, which demonstrates the effectiveness of our proposed method. Specifically, ASR can significantly regularize the model in terms of the inference steps and achieves the accuracy up to $96\%$ for all the three settings. 
It is worth noting that introducing a proper regularization will not affect the ELBO which is the objective function of AIR and AIR-pPrior. The main reason is that ASR can help the model escape from the local optima which violate the structural constraints.

During the training process, all of the three models suffer from sever instability. It results the fact that the nELBO is of large variance. The results largely depend on the initialization and the randomness in the training process. We try to reduce the effect of randomness by fixing the initialization and averaging our results over multiple runs.

\begin{table}[htbp]
    \centering
    \vspace{-.5cm}
    \small{
    \setlength{\tabcolsep}{3pt}
    \caption{Experimental Results on regularization over overlap. Results are averaged over 3 runs.}
    \begin{tabular}{|l|ccc|ccc|}
        \hline
        & \multicolumn{3}{c|}{\small{\textbf{multi-MNIST}}} & \multicolumn{3}{c|}{\small{\textbf{multi-dSprites}}} \\
        Methods      & \multicolumn{1}{c}{nELBO}  & \multicolumn{1}{c}{SE} & \multicolumn{1}{c|}{mIoU} & \multicolumn{1}{c}{nELBO} & \multicolumn{1}{c}{SE} & \multicolumn{1}{c|}{mIoU} \\
        \hline
        AIR   &   $328.5\pm17.1$     &   $37.5\pm 3.8$    &     $0.25\pm 0.03$  &  $341.5\pm 76.5$     & $34.8\pm 8.9$    &  $0.13\pm 0.05$  \\
        AIR-pPrior   &  $\mathbf{306.6}\pm 58.8$     &   $41.5\pm 15.4$    &  $0.35\pm 0.10$     &   $274.3\pm 64.4$    &  $29.3\pm 12.1$     & $0.21\pm 0.13$  \\
        AIR-ASR             &   $337.3\pm 55.1$    &      $\mathbf{36.5}\pm 3.9$ &   $\mathbf{0.67}\pm 0.05$    & $\mathbf{271.8}\pm 18.8$      &    $\mathbf{20.9}\pm 2.1$   & $\mathbf{0.61}\pm 0.03$ \\
        \hline
    \end{tabular}%
    \label{table:overlap}}
    \vspace{-.5cm}
\end{table}%

\vspace{-.1cm}
\subsection{ASR regularization on the overlap}
\vspace{-.1cm}

\begin{figure}
    \vspace{-.1cm}
    \centering
    \begin{subfigure}[t]{0.3\textwidth}
        \includegraphics[width=\textwidth]{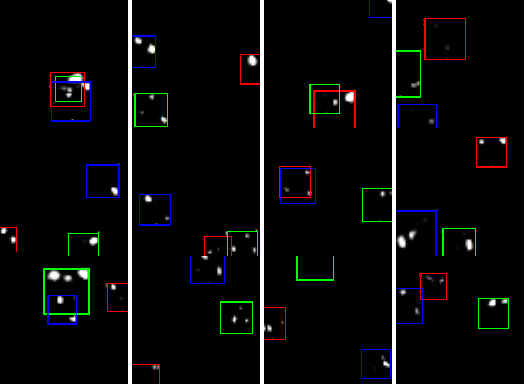}
        \caption{The generative results of AIR-3.}
    \end{subfigure}
    \hfill
    \begin{subfigure}[t]{0.3\textwidth}
        \includegraphics[width=\textwidth]{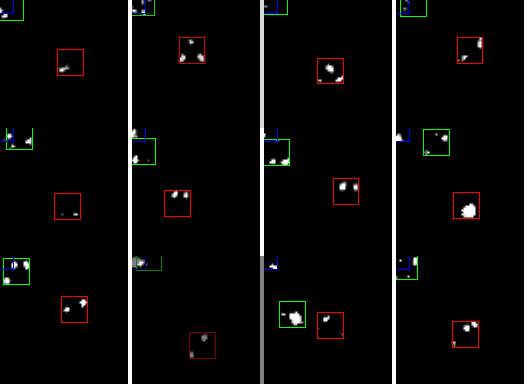}
        \caption{The generative results of AIR-pPrior-3.}
    \end{subfigure}
    \hfill
    \begin{subfigure}[t]{0.3\textwidth}
        \includegraphics[width=\textwidth]{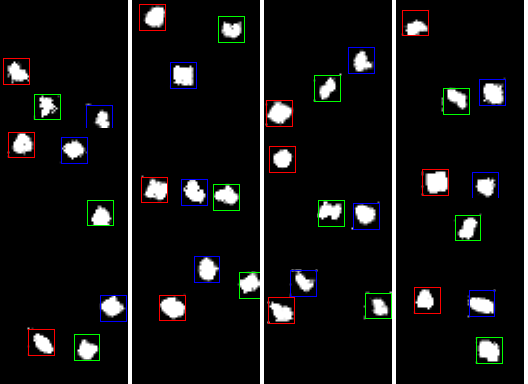}
        \caption{The generative results of AIR-ASR-3.}
    \end{subfigure}
    \vspace{-.2cm}
    \caption{The generative results of Multi-dSprites on 3 objects without overlapping.}
    \label{fig:gen_sprites}
    \vspace{-.6cm}
\end{figure}

When regularizing the overlap, we evaluate models on both Multi-MNIST and Multi-dSprites data. We use 20000 images with three non-overlapping objects as training data and use 1000 images to evaluate performance. Since the number of objects is fixed, we simply set both the generative and inference steps to 3 for fair comparison. We search the hyper-parameters $\lambda_{i=1:7}^{o}$ in $\{1, 10, 20, 100\}$, and we set $\lambda_1^{o}\sim\lambda_5^{o}$ to 1, $\lambda_6^{o}$ to 20, and $\lambda_7^{o}$ to 10.

The reconstruction of Multi-MNIST and generative results of Multi-Sprites are demonstrated in \fig{overlap_mnist_inf} and \fig{gen_sprites} correspondingly. In \fig{overlap_mnist_inf}, the original AIR still merges two objects as one, and it cannot capture the non-overlapping structures. AIR-pPrior has a similar performance. In contrast, AIR-ASR significantly outperforms its baselines, and infers the location of bounding boxes without overlapping. In terms of generative results, the sample quality of AIR-ASR surpasses AIR's and AIR-pPrior's, where the AIR-ASR can generate multiple objects without overlapping whereas its baseline cannot. It demonstrates that the ASR can embed human knowledge into DGMs.

\tabl{overlap} presents the quantitative results. The AIR-ASR surpasses its baselines significantly in terms of mIoU, which indicates that DGMs successfully captures the non-overlapping structures with ASR. It is worth noting that for the Multi-MNIST setting, the nELBO of AIR-pPrior is better than AIR-ASR's. However, AIR-ASR still surpasses AIR-pPrior in terms of the SE and the mIoU, which indicates that AIR-ASR gives better reconstruction results and identifies the location of objects more accurately. This results also verify the claim that simply optimizing the marginal log-likelihood cannot guarantee the generative model to capture the underlying distribution.

\vspace{-.3cm}
\section{Conclusion}
\vspace{-.3cm}

We present a framework ASR to embed human knowledge to improve the inference and generative performance in DGMs. ASR encode human knowledge as a set of structural constraints, and the framework can be optimized efficiently. We use the number of objects and the spatial relationship among them as two examples to demonstrate the effectiveness of our proposed methods. In Multi-MNIST and Multi-dSprites dataset, ASR significantly improves its baselines and successfully captures the underlying structures of the training data.  

In this paper, we mainly focus on the synthetic data, because the AIR cannot deal with complicated images with background. Recently, significant progress has been made in structured generative models~\cite{greff2019multi,burgess2019monet}. ASR can directly be applied to these models, and we left it as future work.

\bibliography{main}

\newpage 
\appendix

\begin{algorithm}[!t]
	\caption{Stochastic Gradient Ascent Training of ASR}
	\label{alg:method}
	\begin{algorithmic}
		\STATE {\bfseries Input:} data $x$, maximum steps $K$, learning rate $\eta$, penalty $R$.
		\STATE Initialize parameters $\theta_0$ and $\phi_0$, and $n=1$.
		\REPEAT
		\STATE $z_{pres}^0 \leftarrow 1$, $z^0 \leftarrow \mathbf{0}$, $t\leftarrow0$
		\REPEAT
		\STATE Update $t \leftarrow t+1$
		\STATE Sample $z_{pres}^t\sim q(z_{pres}^t|z^{<t}, x)$
		\STATE Break if $z_{pres}^i = 0$
		\STATE Sample $z^t\sim q(z^t|z^{<t}, x)$
		\UNTIL $t = K$
		\STATE Update $z\leftarrow (z^1,\cdots, z^t)$
		\STATE Update $KL \leftarrow \log\frac{q(z)}{p(z)}$, $Rec \leftarrow \log p(x|z)$, $r \leftarrow R(q)$
		\STATE Update $J^\prime(\theta, \phi) \leftarrow Rec-KL-r$
		\STATE Update $\theta$ and $\phi$: $\theta_{n} \leftarrow \theta_{n-1} + \eta\frac{\partial J^\prime(\theta, \phi)}{\partial\theta}$, $\phi_{n} \leftarrow \phi_{n-1} + \eta\frac{\partial J^\prime(\theta, \phi)}{\partial\phi}$
		\STATE Update $n$: $n\leftarrow n+1$
		\UNTIL{Both $\theta$ and $\phi$ converge.}
	\end{algorithmic}
\end{algorithm}

\section{Algorithm}

The training algorithm of ASR is described in Algorithm~\ref{alg:method}.

\section{The reconstruction results}
In this section, we illustrates all the inference results of AIR, AIR-pPrior and AIR-ASR on all settings. Results can be found in Fig.1 $\sim$~Fig.5. Some of them are included in the main body.

\begin{figure}[bh]
	\vspace{-.2cm}
	\centering
	\begin{subfigure}[t]{0.3\textwidth}
		\includegraphics[width=\textwidth]{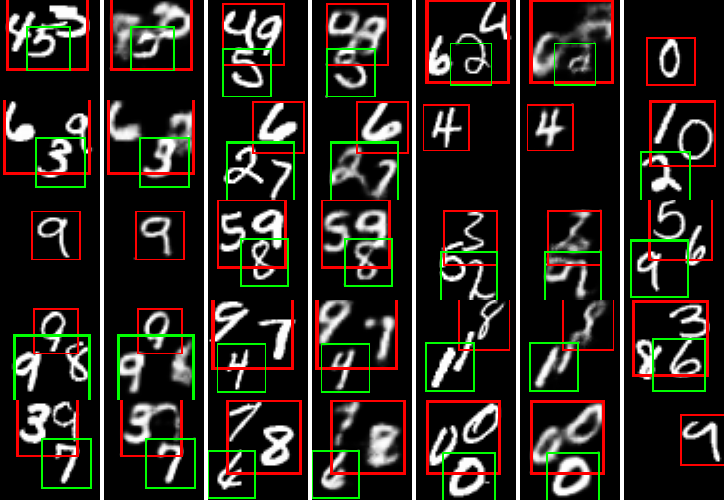}
		\caption{The reconstruction results of AIR-13.}
	\end{subfigure}
	\hfill
	\begin{subfigure}[t]{0.3\textwidth}
		\includegraphics[width=\textwidth]{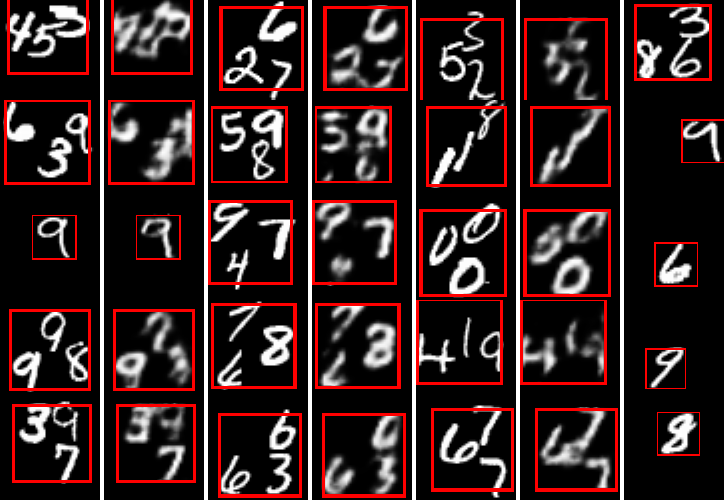}
		\caption{The reconstruction results of AIR-pPrior-13.}
	\end{subfigure}
	\hfill
	\begin{subfigure}[t]{0.3\textwidth}
		\includegraphics[width=\textwidth]{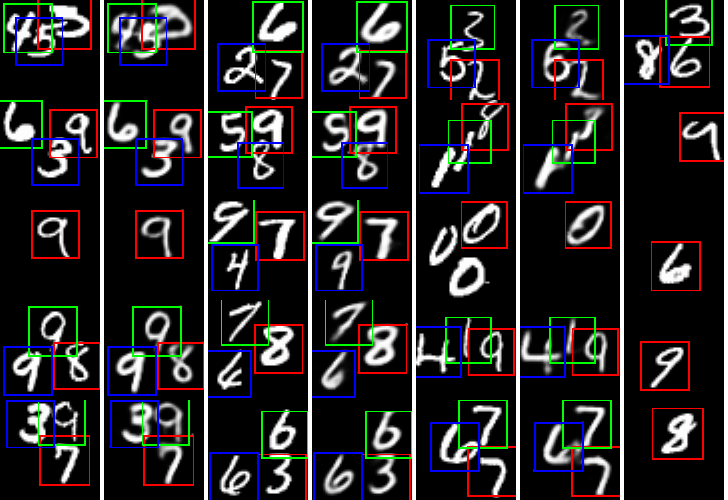}
		\caption{The reconstruction results of AIR-ASR-13.}
	\end{subfigure}
	\caption{The reconstruction results of Multi-MNIST on 1 or 3 objects.}
	\label{fig:num-13}
\end{figure}

\begin{figure}[bh]
	\vspace{-.2cm}
	\centering
	\begin{subfigure}[t]{0.3\textwidth}
		\includegraphics[width=\textwidth]{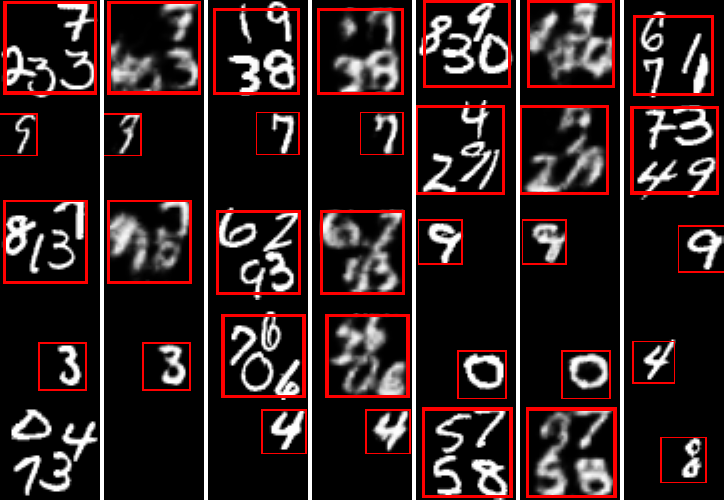}
		\caption{The reconstruction results of AIR-14.}
	\end{subfigure}
	\hfill
	\begin{subfigure}[t]{0.3\textwidth}
		\includegraphics[width=\textwidth]{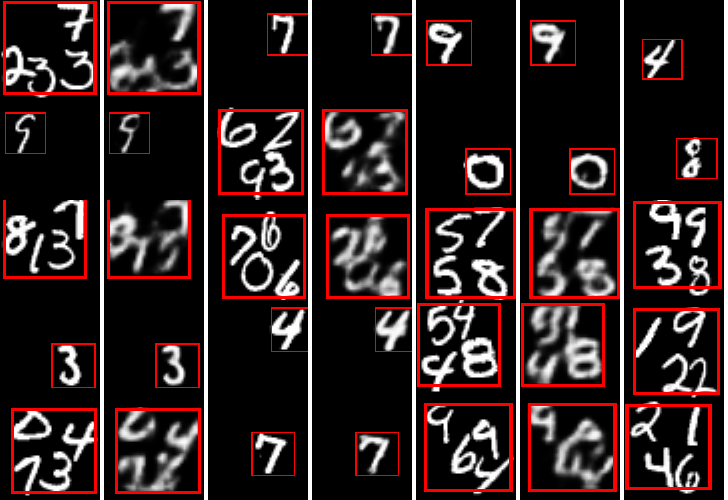}
		\caption{The reconstruction results of AIR-pPrior-14.}
	\end{subfigure}
	\hfill
	\begin{subfigure}[t]{0.3\textwidth}
		\includegraphics[width=\textwidth]{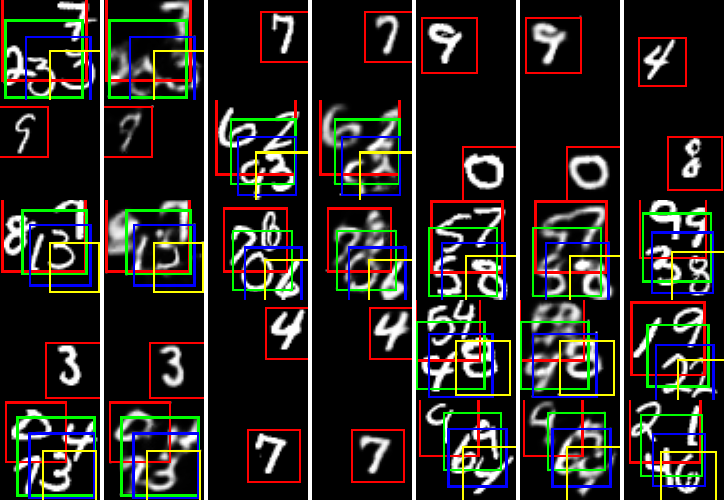}
		\caption{The reconstruction results of AIR-ASR-14.}
	\end{subfigure}
	\caption{The reconstruction results of Multi-MNIST on 1 or 4 objects.}
	\label{fig:num-14}
\end{figure}

\begin{figure}[bh]
	\vspace{-.2cm}
	\centering
	\begin{subfigure}[t]{0.3\textwidth}
		\includegraphics[width=\textwidth]{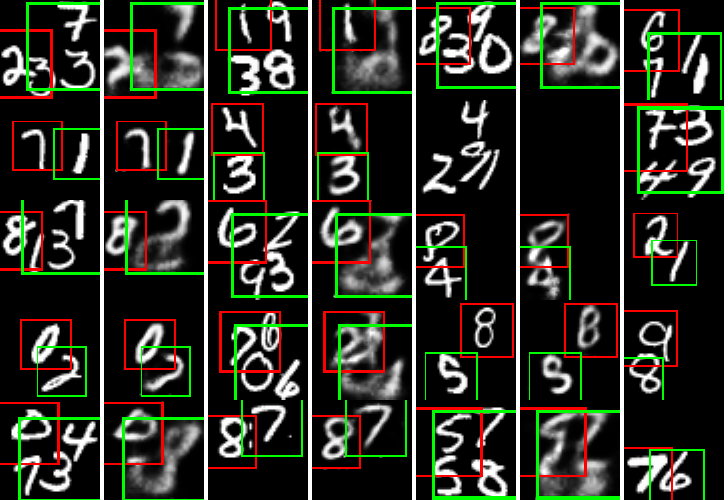}
		\caption{The reconstruction results of AIR-24.}
	\end{subfigure}
	\hfill
	\begin{subfigure}[t]{0.3\textwidth}
		\includegraphics[width=\textwidth]{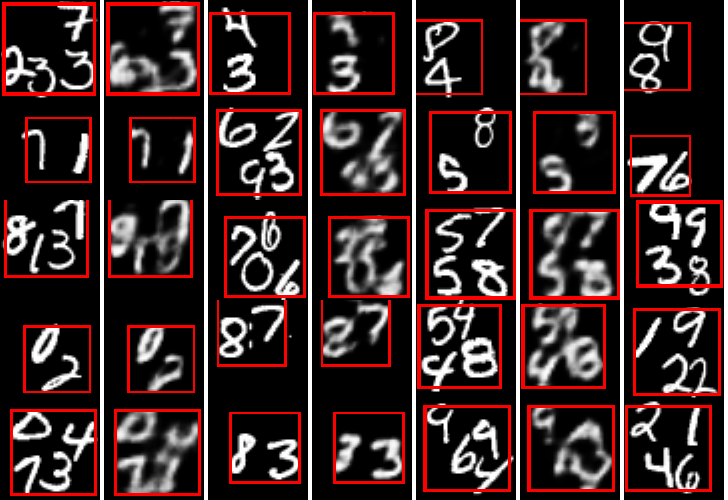}
		\caption{The reconstruction results of AIR-pPrior-24.}
	\end{subfigure}
	\hfill
	\begin{subfigure}[t]{0.3\textwidth}
		\includegraphics[width=\textwidth]{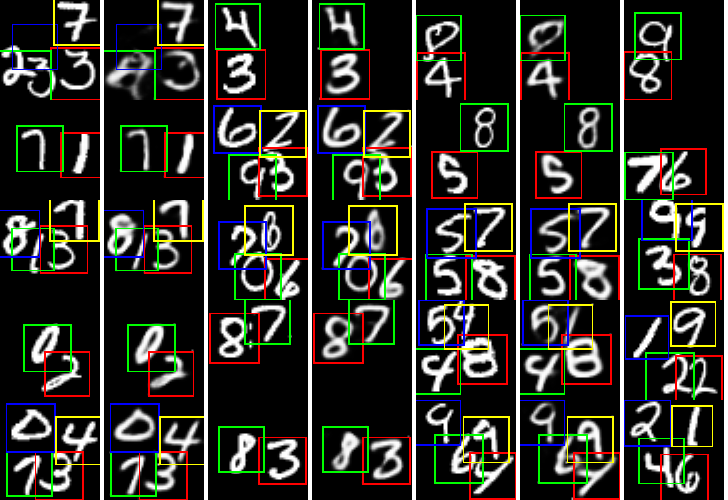}
		\caption{The reconstruction results of AIR-ASR-24.}
	\end{subfigure}
	\caption{The reconstruction results of Multi-MNIST on 2 or 4 objects.}
	\label{fig:num-24}
\end{figure}

\begin{figure}[bh]
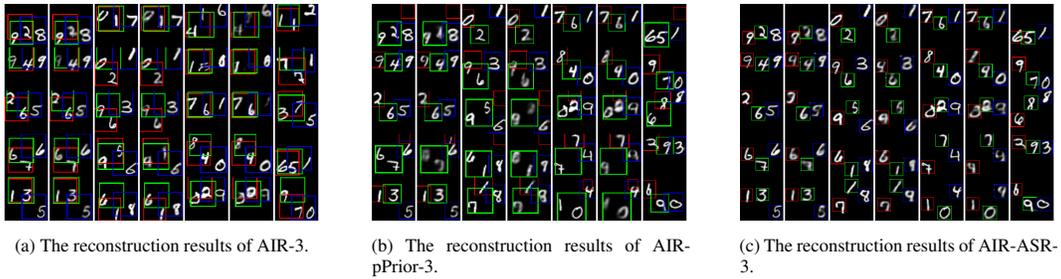

	\vspace{-.2cm}
	\centering
	\begin{subfigure}[t]{0.3\textwidth}
		\includegraphics[width=\textwidth]{nAIR-3.png}
		\caption{The reconstruction results of AIR-3.}
	\end{subfigure}
	\hfill
	\begin{subfigure}[t]{0.3\textwidth}
		\includegraphics[width=\textwidth]{nAIR-pPrior-3.png}
		\caption{The reconstruction results of AIR-pPrior-3.}
	\end{subfigure}
	\hfill
	\begin{subfigure}[t]{0.3\textwidth}
		\includegraphics[width=\textwidth]{nAIR-ASR-3.png}
		\caption{The reconstruction results of AIR-ASR-3.}
	\end{subfigure}
	\caption{The reconstruction results of Multi-MNIST 3 objects without overlapping.}
	\label{fig:num-3}
\end{figure}

\begin{figure}[bh]
	\vspace{-.2cm}
	\centering
	\begin{subfigure}[t]{0.3\textwidth}
		\includegraphics[width=\textwidth]{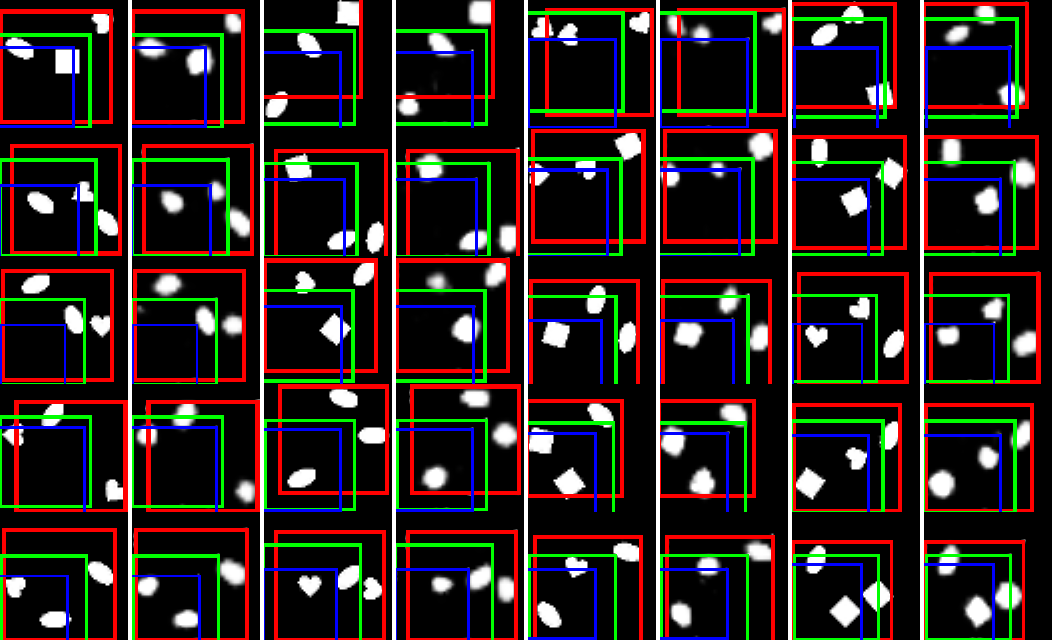}
		\caption{The reconstruction results of AIR-3.}
	\end{subfigure}
	\hfill
	\begin{subfigure}[t]{0.3\textwidth}
		\includegraphics[width=\textwidth]{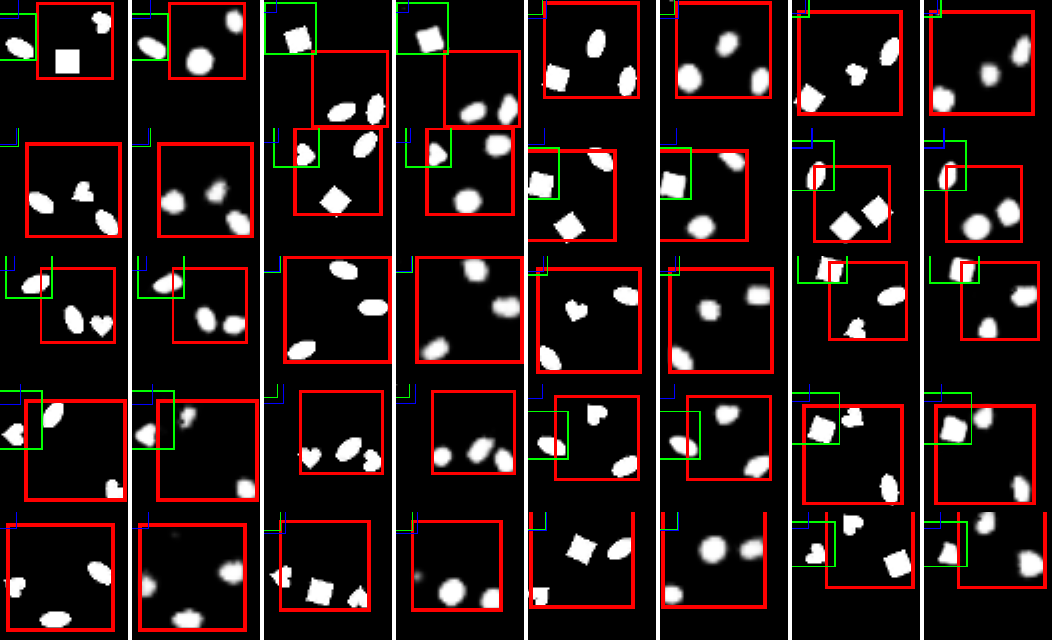}
		\caption{The reconstruction results of AIR-pPrior-3.}
	\end{subfigure}
	\hfill
	\begin{subfigure}[t]{0.3\textwidth}
		\includegraphics[width=\textwidth]{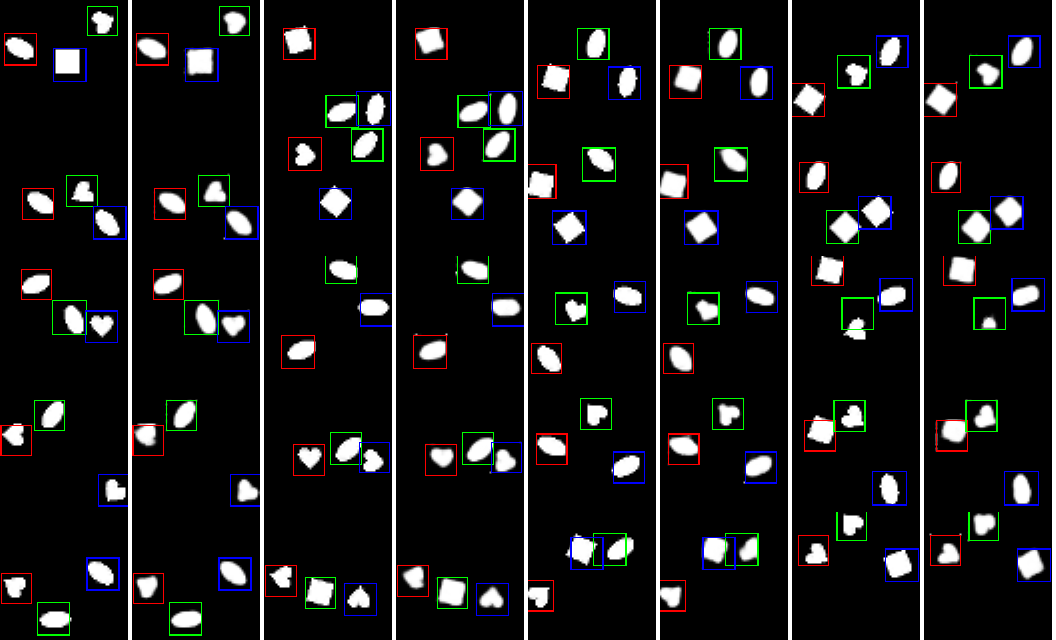}
		\caption{The reconstruction results of AIR-ASR-3.}
	\end{subfigure}
	\caption{The reconstruction results of Multi-Sprites 3 objects without overlapping.}
	\label{fig:num-3-sprites}
\end{figure}

\end{document}